\documentclass{tADR2e}

\usepackage[colorlinks=false]{hyperref}
\usepackage{color}
\graphicspath{{figures/}}
\usepackage{amsmath}

\begin{document}

\articletype{}

\title{High-level Features for Resource Economy and Fast Learning in Skill Transfer}

\author{Alper Ahmetoglu$^{a\ast}$\thanks{${^\ast}$Corresponding author. Email: alper.ahmetoglu@boun.edu.tr
\vspace{6pt}}, Emre Ugur$^{a}$, Minoru Asada$^{b}$, Erhan Oztop$^{b, c}$\\
\vspace{6pt}
$^{a}${\em{Department of Computer Engineering, Bogazici University, Turkey}}\\
$^{b}${\em{OTRI/SISREC, Osaka University, Osaka, Japan}}\\
$^{c}${\em{Department of Computer Science, Ozyegin University, Turkey}}\\
}

\maketitle

\begin{abstract}
Abstraction is an important aspect of intelligence which enables agents to construct robust representations for effective and efficient decision making. In the last decade, deep neural networks are proven to be effective learning systems that are applicable to a wide range of application domains, in particular, due to their ability to form increasingly complex abstractions at successive layers in an end-to-end fashion. However, these abstractions, unless enforced, are mostly distributed over many neurons, making the re-use of a learned skill costly and blind to the insights that can be obtained on the emergent representations. Previous work either enforced formation of abstractions creating a designer bias, or used a large population of neural units without investigating  how to obtain high-level low-dimensional features that may more effectively capture the source task. For avoiding designer bias and unsparing resource use, we propose to exploit neural response dynamics to form compact representations to use in skill transfer. For this, we consider two competing methods based on (1) information loss/maximum information compression principle and (2) the notion that abstract events tend to generate slowly changing signals, and apply them to the neural signals generated during task execution. To be concrete, in our simulation experiments, we either apply principal component analysis (PCA) or slow feature analysis (SFA) on the signals collected from the last hidden layer of a deep neural network while it performs a source task, and use these features for skill transfer in a new, target, task. We then compare the generalization and learning performance of these alternatives with the baselines of skill transfer with full layer output and no-transfer settings. Our experimental results on a simulated tabletop robot arm navigation task show that units that are created with SFA are the most successful for skill transfer. SFA as well as PCA, incur less resources compared to usual skill transfer where full layer outputs are used in the new task learning, whereby many units formed show a localized response reflecting end-effector-obstacle-goal relations. Finally, SFA units with lowest eigenvalues resembles symbolic representations that highly correlate with high-level features such as joint angles and end-effector position which might be thought of precursors for fully symbolic systems.
\medskip
\end{abstract}

\medskip

\section{Introduction}
\label{sec:introduction}
It can be argued that intelligent behavior, both artificial and biological agents, require the use of abstractions for high competency in the real world \cite{Konidaris2019}. Abstraction allows an agent to filter out irrelevant aspects of incoming stimuli, and thus, equips it with a robust representation. From a biological perspective, it is not clear how concept and symbol representations are formed by the central nervous system \cite{Taniguchi2019}. Yet, behavioral markers of such representations can be seen even in animals at the lower levels of cognitive hierarchy. For example, toads have abstract prey and predator feature detectors that respond according to the high level features present in the visual stimuli related to the size and shape relative to the direction of motion \cite{Ewert2004book}. From the robotics point of view, classically, a set of concepts or symbols can be defined apriori and later grounded to the sensorimotor stimuli of the agent acting in its environment \citep{Petrick2008}. More recent approaches aim to abstract representations directly from the sensorimotor experience of the agent \citep{Sun2000, Konidaris2014, asai2018classical, ahmetoglu2020deepsym}.

In fact, the deep learning approaches form complex feature representations by combining simpler ones in successive layers \cite{bengio2009learning}. As deep neural networks have been very successful in many real world problems such as image recognition \cite{krizhevsky2012imagenet}, machine translation \cite{sutskever2014sequence}, playing the game of Go \cite{silver2016mastering}, and protein folding problem \cite{senior2020improved}, most of the focus has been put on improving performance and finding newer applications of deep learning. One of the bottlenecks of learning in deep networks is the large amount of data required for learning tasks from scratch. To attack this problem, knowledge or skill transfer methods allowing the reuse of learned representations have started to appear \cite{weiss2016survey, vanschoren2018meta}, although similar ideas were present earlier in the robot learning domain \citep{ThrunTom2016}.
One straightforward way to achieve knowledge transfer is to create a new network with access to all the layers of original network that was used to learned the initial task (e.g. \cite{Rusu2016progressive, LiDerek2016}). 
This ensures that any representation discovered will have a chance of exploitation during new task learning. However, in these approaches all the units in the layers are linked to the new network through adaptable weights, and thus, which of those units need to be exploited for the new task is left for the learning algorithm to find. Considering resource economy both in terms of memory and energy, this seems wasteful. In this sense, a principled task-agnostic way of selecting \textit{which} units to use for the next task learning is highly desirable.  This study aims to make a step forward in this direction. 

Earlier robotic work indicates that units that capture high level features are potential candidates to facilitate effective knowledge transfer \cite{Ugur-2015-ICRA}. From a neural computation point of view, this also makes sense as compact high-level representations would be more economical to process and pass around in the neural circuits of biological systems. Thus, the brain might adopt a strategy to represent sensorimotor information compactly with minimum information loss, inline with the information compression ideas \cite{Schmidhuber2008,Wolff2016}. Another hypothesis can be obtained by generalizing the idea that slowly changing features in sensory data tend to correspond to more high-level concepts \cite{wiskott2002slow} to neural responses.  According to this view, the brain might seek to exploit those neurons that are more stable over the others that show frequent changes. For example, when a hand waving action is observed, the earlier sites in the visual processing pathway (e.g. areas V1, V2) would show temporally changing activity where as at the end of the processing pipeline (i.e. in area IT), the recognition of the hand waving action would be represented with a few neurons, which would show stable activity during the most of the observation period over a variety hand waving actions.

Motivated from these hypotheses, in this paper, we analyze different ways of transferring previously learned representations for a new task in an economical way. More specifically, we consider two methods: (1) principal component analysis (PCA) for the minimum information loss, and (2) slow feature analysis (SFA) \cite{wiskott2002slow} for creating signals that change slowly. PCA reduces the dimensionality of the data while preserving the information maximally which is a suitable candidate for transferring compact features. On the other hand, SFA, which is not very well explored in the transfer learning context, creates slowly changing representations which are arguably more robust because objects of interest in the real world do not make abrupt changes considering the notion of object permanency. In our experiments, we first train a deep Q-network \cite{mnih2013playing} on a task, where the goal is to move the robot arm to a desired target position in the presence of a rectangular obstacle which may appear at different locations. After training, we create separate skill transfer scenarios in which either PCA or SFA  transformations are applied on the last hidden layer activations of the network and used in new task learning by augmenting the features found to the last hidden layer of the new network. We compare these scenarios with the baselines of skill transfer with full layer output and no-transfer scenarios. Our experimental results show that:
\begin{itemize}
    \item Using features that are constructed with SFA is not only more economical in terms of the number of units, but also better for skill transfer than the naive approach of using the full set of layer activations.
    \item PCA is also helpful for resource economy in skill transfer, but not as good as SFA in terms of success rate of the new task.
    \item SFA and PCA capture interpretable high-level features such as joint angles, tip locations, and the distance from the tip position to the goal position, solely from the activation history of the network.
\end{itemize}
In the rest of this paper, we detail our methodology in Section \ref{sec:methods}, define the experimental setup in Section \ref{sec:exp_setup}, give the results in Section \ref{sec:results}, and conclude in Section \ref{sec:conclusion}.

\subsection{Related Work}
There has been an increased attention on learning high-level abstractions from the experience of the robot. Konidaris et al. proved that learning symbols for the pre- and post-conditions of executed actions are necessary and sufficient for determining the feasability of a plan \cite{Konidaris2014}. Based on this, they constructed symbols for pre-conditions and effects of executed actions in a 2d game environment. In a follow-up work, they moved this framework into the probabilistic setting and used the learned symbols in a real-world robot to make plans \cite{konidaris2018skills}. Ugur and Piater followed an object-centric strategy and created symbols from clusters of object features and effects for each action \cite{Ugur-2015-ICRA}. A more recent follow-up work uses deep neural networks with a binary bottleneck layer to predict the effect of executed actions in an end-to-end fashion \cite{ahmetoglu2020deepsym}. Activations in the bottleneck layer are then treated as symbols and used for planning. A similar work also employs deep neural networks with a binary layer but symbols are learned independently from the executed actions and observed effects. All of these works force symbols to be discrete and this effectively limits the capability of the system. A more desirable property would be to allow the system to use continuous representations and observe discrete symbols only as a by-product of some learning process. To this end, in this paper, we explore different methods of constructing features in a generic, task-agnostic way to investigate whether they exhibit a symbol-like meaning.

PCA is a frequently used method for dimensionality reduction and data analysis due to its ability to create features that capture the data with a desired level of information loss. In particular, it allows the tuning of the amount of information loss by choosing the principal vectors to represent data by simply examining the eigenvalues corresponding to them. Santello et al. used PCA to analyze the intrinsic dimensionality of hand posture for grasping different objects and found that the first two dimensions cover more than 80\% of the variance \cite{santello1998postural}. Tripathi and Wagatsuma applied PCA to different joint sets and time segments to impose a prior on the PCA algorithm based on the control approach of the central nervous system \cite{tripathi2016pca}. Promsri and Federolf analyzed the effect of using PCA on the acceleration of postural changes as opposed using PCA only on the amplitude \cite{promsri2020analysis}. As a non-linear approach, Chen et al. combined sparse autoencoders with dynamic movement primitives \cite{schaal2006dynamic} for 50-dimensional human movement data, and showed that new data can be generated using only a single neuron of the trained system \cite{chen2015efficient}.

SFA \cite{wiskott2002slow} can be loosely thought as the application of PCA to the first time-derivative of the data. Therefore, it is particularly useful for extracting slowly changing signals when the low eigenvalue vectors are chosen. Dawood and Loo used an incremental version of SFA \cite{kompella2011incremental} for action segmentation from high-dimensional video input. A similar work to ours studied the use of hierarchical SFA features extracted from raw image data in the context of reinforcement learning \cite{legenstein2010reinforcement}. Our study differs from this work as we are interested in extracting high-level features from an already formed network, be it either prewired or learned, and use them in subsequent tasks.

\section{Methods}
\label{sec:methods}
Suppose that an agent has developed a neural system to solve a specific task. When the agent encounters a similar task, neurons in this system will respond to the sensorimotor input albeit possibly in a different way since the input will be different. Even though the system does not directly provide the appropriate motor control output, it would be economically viable to use a previously learned network instead of re-learning everything from scratch. We focus on extracting features from this network in an economical way. Note that it is not important
how this network has been formed; it might have been learned via stochastic gradient descent, pre-wired, or obtained through an evolutionary algorithm.

Firstly, we train a deep Q-learning agent \cite{mnih2013playing} on a primary task, so to have a network which we can extract features from. In the primary task, a robot arm tries to move its arm to the goal position while avoiding a rectangular obstacle. After training, we extract features from this network in three different ways and use them in a secondary task to assess the bootstrapping effect induced for the new task learning. As a baseline we also learn the new task from scratch without any transfer. So, overall, we have these cases:
\begin{enumerate}
    \item \textbf{Learning from scratch (\texttt{transfer:none}).} The new task is trained without any transfer to form a baseline for the next three transfer scenarios.
    \item \textbf{The last hidden layer activations (\texttt{transfer:full}).} This is one of the most frequently used methods for transferring visual features from networks that are trained on large-scale data sets \cite{goodfellow2016deep} and serves as a reference for the next two methods.
    \item \textbf{Transforming the last hidden layer with PCA (\texttt{transfer:PCA}).} While the last layer activations provide useful information, it is mostly distributed in many units. PCA is especially useful for concentrating the distributed information into fewer units, effectively reducing the number of neurons, thus reducing the number of weights that needs to be optimized for learning the new task. Let us denote our dataset as $X=\{H_i\}_{i=1}^N$ where $N$ is the number of robot movement trajectories arising from executing the primary task of the robot, and $H_i$ is an $T_i \times D$ matrix where $T_i$ is the number of timesteps in $i$th trajectory and $D$ is the dimensionality of the last hidden layer. We concatenate each $H_i$ from the first dimension so that $X$ is a $(T_1+T_2+\dots+T_N) \times D$ matrix. We want to find a projection $z=Xw$ such that $\text{Var}(z)$ is maximized. This leads to the following objective:
    \begin{equation}
        \arg \max_{w} \quad \frac{w^TX^TXw}{w^Tw}
    \end{equation}
    which is the Rayleigh quotient, and the expression takes its maximum value when $w$ is equal to the eigenvector with the largest eigenvalue of $X^TX$ \cite{trefethen1997numerical}.

    \item \textbf{Transforming the last hidden layer with SFA (\texttt{transfer:SFA}).} PCA puts an emphasis on the maximum information preserving units. On the other hand, SFA tries to minimize the time-derivative of the output signals. More specifically, SFA creates output features $z=Xw$ with the following objective and restrictions \cite{wiskott2002slow}:
    \begin{equation}
        \min \quad \mathbb{E}[\dot{z}_i^2]
    \end{equation}
    subject to
    \begin{align}
    \mathbb{E}[z_i] &= 0 \label{eq:zeromean}\\
    \mathbb{E}[z_i^2] &= 1 \label{eq:unitvar}\\
    \mathbb{E}[z_iz_j] &= 0 \quad \forall j < i \label{eq:decorr}
    \end{align}
    Here, Equations \ref{eq:zeromean} and \ref{eq:unitvar} prevents the trivial solution of a constant feature and also forces the output to be normalized. Equation \ref{eq:decorr} forces features to be orthogonal to each other. The solution can be found by first whitening the data, and then applying PCA to the time derivative of $X$. This method is shown to be useful for extracting the independent factors of an input signal.
\end{enumerate}

\begin{figure}[htbp]
    \centering
    \subfigure[\texttt{transfer:full} \label{subfig:transfer_full}]{
        \includegraphics[width=0.4\textwidth]{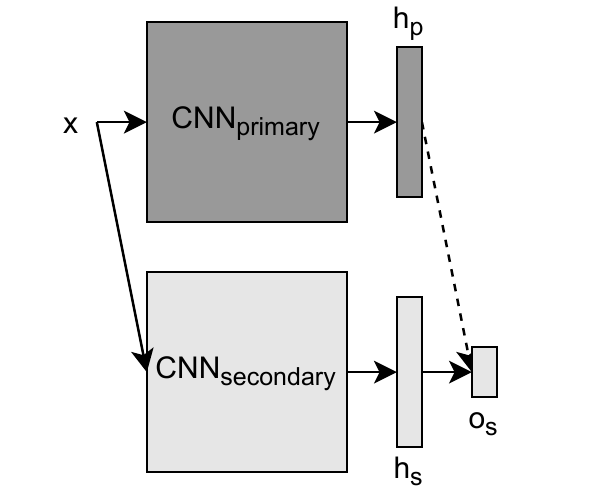}
    }
    \subfigure[\texttt{transfer:sfa} \label{subfig:transfer_sfa}]{
        \includegraphics[width=0.4\textwidth]{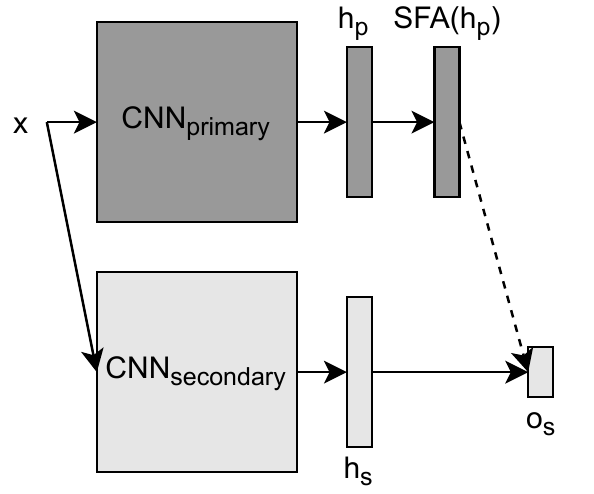}
    }
    \caption{Different transfer methods are depicted. \texttt{transfer:pca} is identical to \ref{subfig:transfer_sfa} except that the transformation is done with PCA, instead of SFA.}
    \label{fig:transfer}
\end{figure}

We transfer the information from a network with these three methods to a new task by concatenating the features to the input of the last layer as shown if Figure \ref{fig:transfer}. In our experiments, we define the secondary task to be a reinforcement learning setup as well. The environment is similar to the primary task except that there is a different type of obstacle to avoid.

Note that both PCA and SFA defines an affine transformation of the last layer, and thus, does not bring any advantage in terms of function complexity. However, re-alignment of the feature space can greatly accelerate the speed of convergence with stochastic gradient descent \cite{lecun2012efficient}. Likewise, we expect that agents that use these transformations will exhibit a better performance with fewer samples.

\section{Experiment Setup}
\label{sec:exp_setup}

\begin{figure}[htbp]
    \centering
    \includegraphics[width=0.75\textwidth]{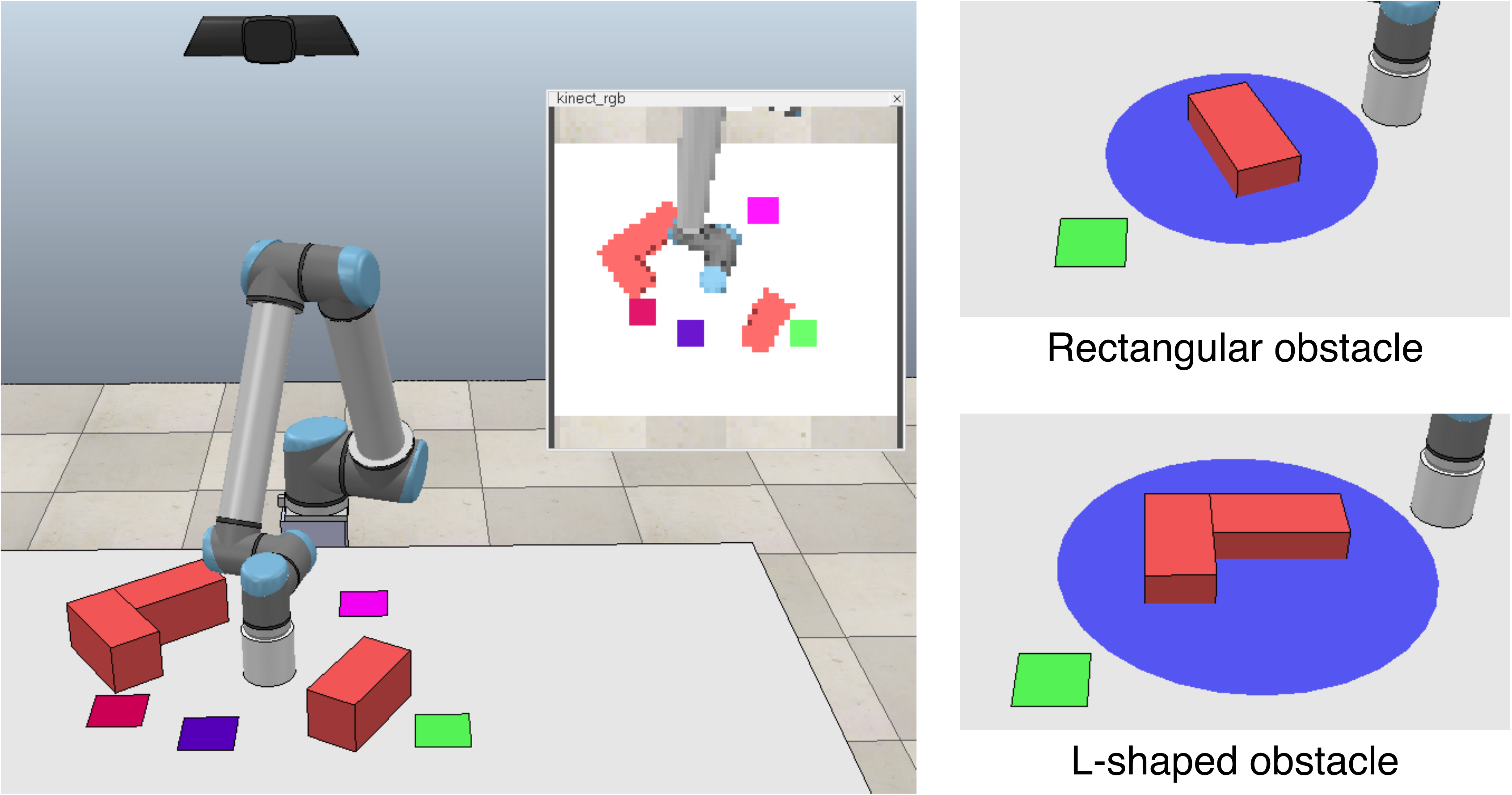}
    \caption{Left: The experiment setup. The obstacles are red-colored, and the goal marker is green-colored. Distractors dynamically change their colors in the RGB range of [(0, 0, 0)-(255, 0, 255)]. The corresponding input from the camera is shown on the top-right inset. Right: The regions of penalty around obstacles  are visualized in blue (see text).}
    \label{fig:setup}
\end{figure}

We perform our experiments on CoppeliaSim 4.2.0 simulator \cite{coppeliaSim}. The experiment setup consists of a UR10 \cite{ur10} robot arm with a solid cylinder attached to its end effector and a camera placed at 180cm over the table for a top-down visual perception. The robot arm can make planar movements in 8 different uniform directions in a $80$cm $\times$ $100$cm rectangular workspace. The camera input provides a 64 $\times$ 64 pixels colored image at each timestep (Figure \ref{fig:setup}). There are rectangular shaped and L-shaped obstacles, a goal marker, and distractor markers in the environment. The general overview of the setup is shown in Figure \ref{fig:setup}.

In this environment, we define two different tasks, a primary task in which the robot tries to move its end-effector to the goal position while avoiding a rectangular obstacle, and a secondary task where there is an L-shaped obstacle instead of a rectangular one to avoid. In both tasks, we included distractors that change colors and move in random directions with a constant speed to increase the difficulty of the tasks. The reward function for both environments is defined as follows:
\begin{align}
    R(t) =
    \begin{cases}
    \hfil 10 \quad &\text{if} \quad \|x_{\text{tip}}(t) - x_{\text{goal}}\|_2 < 5\text{cm}\\
    \hfil 10 \Delta_{t}(x_{\text{tip}}, x_{\text{goal}}) \quad &\text{else if} \quad \|x_{\text{tip}}(t) - x_{\text{obstacle}}(t)\|_2 > \tau\\
    \hfil 10 (\Delta_{t}(x_{\text{tip}}, x_{\text{goal}}) + \text{relu}(\Delta_{t}(x_{\text{tip}}, x_{\text{obstacle}}))) \quad &\text{otherwise}
    \end{cases}
    \label{eq:reward}
\end{align}
where $\tau$ is a threshold for obstacle proximity penalty and $\Delta_t$ denotes the change in the distance between robot end-effector and the target over consecutive time steps:
\begin{equation}
    \Delta_t(x, y) = \|x(t-1)-y(t-1)\|_2 - \|x(t)-y(t)\|_2
\end{equation}
We set $\tau$ to 21cm for the primary environment and 28cm for the secondary environment. The robot gets a penalty whenever it goes towards the obstacle when it is in the blue region visualized in Figure \ref{fig:setup}.

We use a convolutional neural network (CNN) architecture, where three convolutional layers followed by two fully-connected layers are employed in each architecture. Convolutional layers have 16, 32, and 64 channels successively. Each layer has a kernel size of $3 \times 3$ with a stride of 2 and a padding of 1. Fully-connected layers have 512 and 9 number of units. We concatenate the transferred features to the last hidden layer except the learning from scratch case. For example, if we transfer 100 slow features, the dimensionality of the last hidden layer  becomes $512+100=612$.

We train each model as a deep Q-network \cite{mnih2013playing} for 2000 episodes with a replay buffer of size 50,000. Each episode lasts for at most 200 time-steps, after then, the episode terminates. We used the reward function defined in Equation \ref{eq:reward}. The last fully-connected layer is used for the Q-value estimation. There are 8 different actions for 8 different directions and an additional action for no operation. For the optimization, we use Adam optimizer \cite{kingma2014adam} with a learning rate of 0.001 with no learning rate decay.

\section{Results}  
\label{sec:results}

In this section, we first test the generalization performance attained by the transfer of features acquired with different methods to a new task in \ref{subsec:transfer}. Next, we analyze correlations between the proposed features and high-level task related features in Section \ref{subsec:correlation}. Lastly, in Section \ref{subsec:visualization}, we visualize the responses of neurons for varying inputs to get an insight about their functionalities.

\subsection{Transfer Performance}
\label{subsec:transfer}
The aim of this experiment is to test the bootstrapping effect of using a previously learned representation on learning a similar task from scratch. To this end, we concatenate the activations of a previously trained network (described in the previous section) to the last hidden layer. This method is one of the basic transfer learning methods \cite{goodfellow2016deep}. We compare the transfer of plain activations, PCA features, and SFA features. In addition, we train a network with no transfer to observe the bootstrapping effect.

\begin{figure}[htbp]
    \centering
    \includegraphics[width=0.6\textwidth]{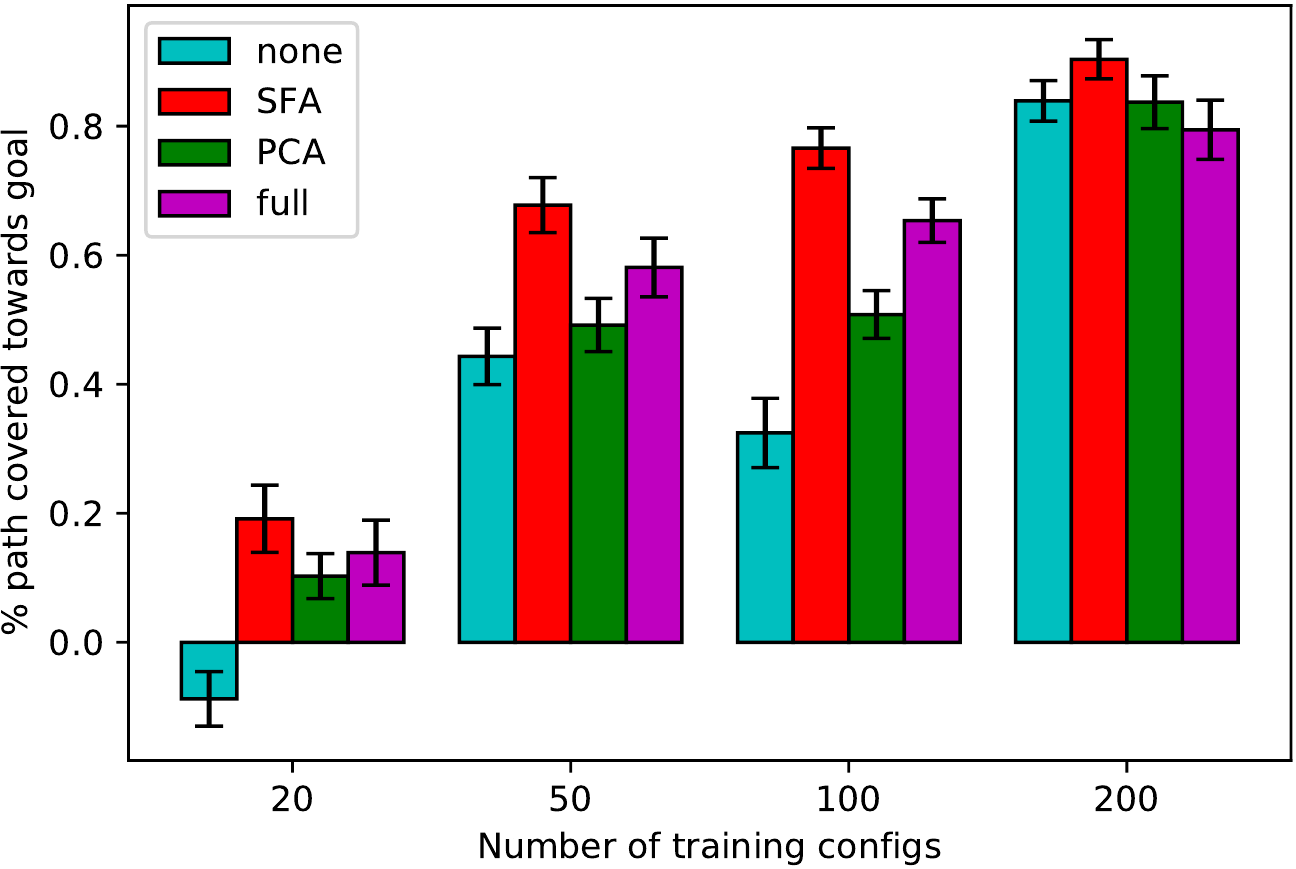}
    \caption{Generalization performance measured as distance covered towards the goal vs. number of configurations experienced during learning in the new task. Welch's t-test \cite{welch1947generalization} shows significant difference between \texttt{transfer:SFA} and \texttt{transfer:full} for 100 and 200 configurations ($p=0.022$ and $p=0.041$, respectively), and almost significant difference for 50 configurations ($p=0.082$).}
    \label{fig:config_rewards}
\end{figure}

Each model is trained with 20, 50, 100, and 200 number of training configurations for 2000 episodes. Here, a training configuration refers to an environment setting (i.e. initial position of the objects). Each episode is initialized with a configuration sampled from this set. After convergence, we test each model at each 500 episodes on the previously unseen environment settings (i.e. configurations) to observe the generalization performance. For testing, we collect 100 runs with each model and calculate the average percentage of path covered towards the goal, and treat this estimate as one test result. The average of 15 different test results at best performing episodes (validated with 5 results) for each model are reported in Figure \ref{fig:config_rewards}.

We observe that using \texttt{transfer:SFA} condition gives better performance compared to \texttt{transfer:PCA}, \texttt{transfer:full} and \texttt{transfer:none} conditions. Note that SFA and PCA features are 100 dimensional vectors while plain activations are 512 dimensional vectors. This result shows that SFA indeed creates more condensed features that are appropriate for skill transfer. The bootstrapping effect fades away when we increase the number of training configurations to 200 configurations. This is an expected result as every model gets better when we increase the variation in the training set. However, the usage of low training configurations is desirable in many real world settings.

\begin{figure}[hbtp]
    \centering
    \subfigure[50 configs \label{subfig:50config}]{\includegraphics[width=0.32\textwidth]{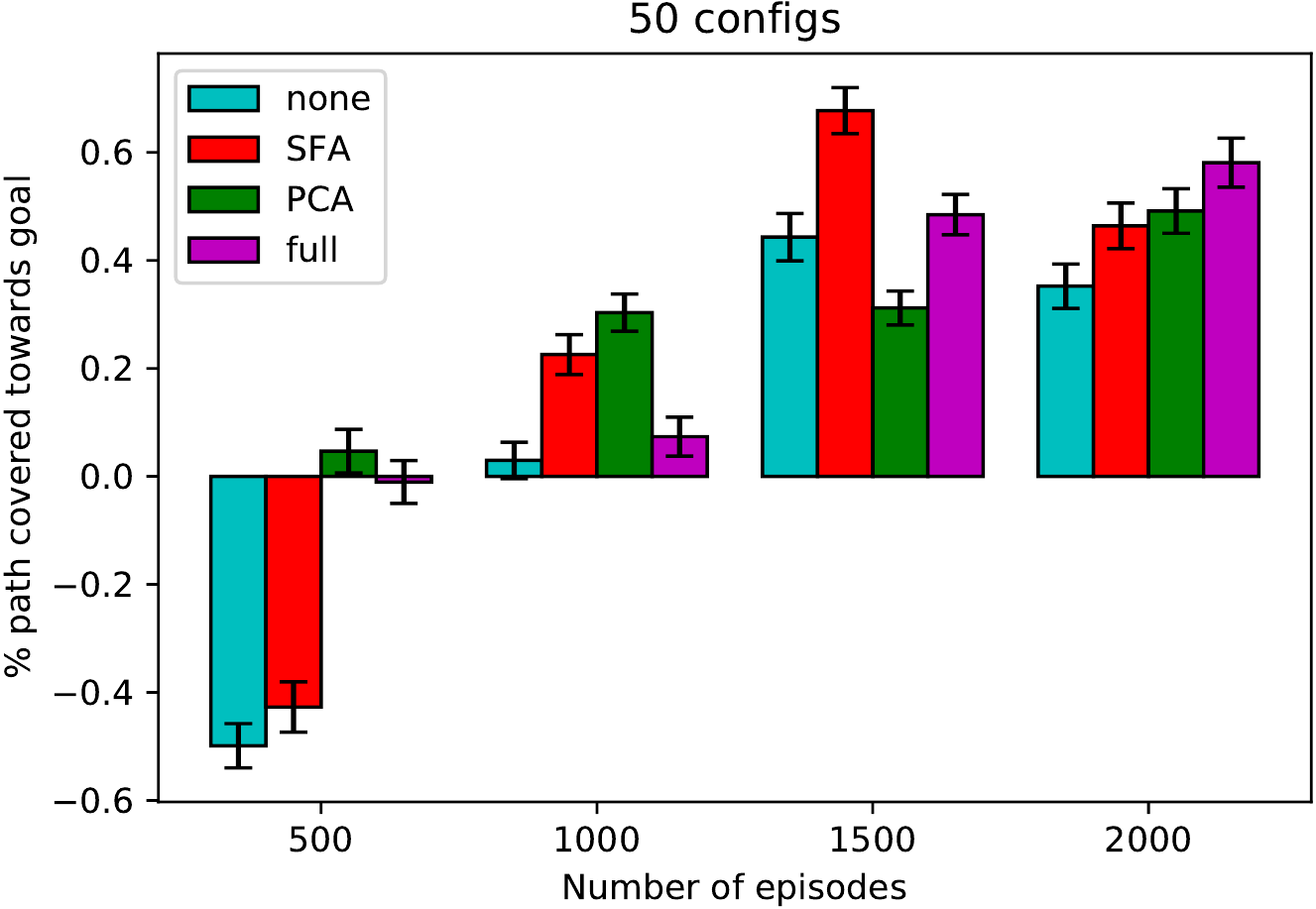}}
    \subfigure[100 configs \label{subfig:100config}]{\includegraphics[width=0.32\textwidth]{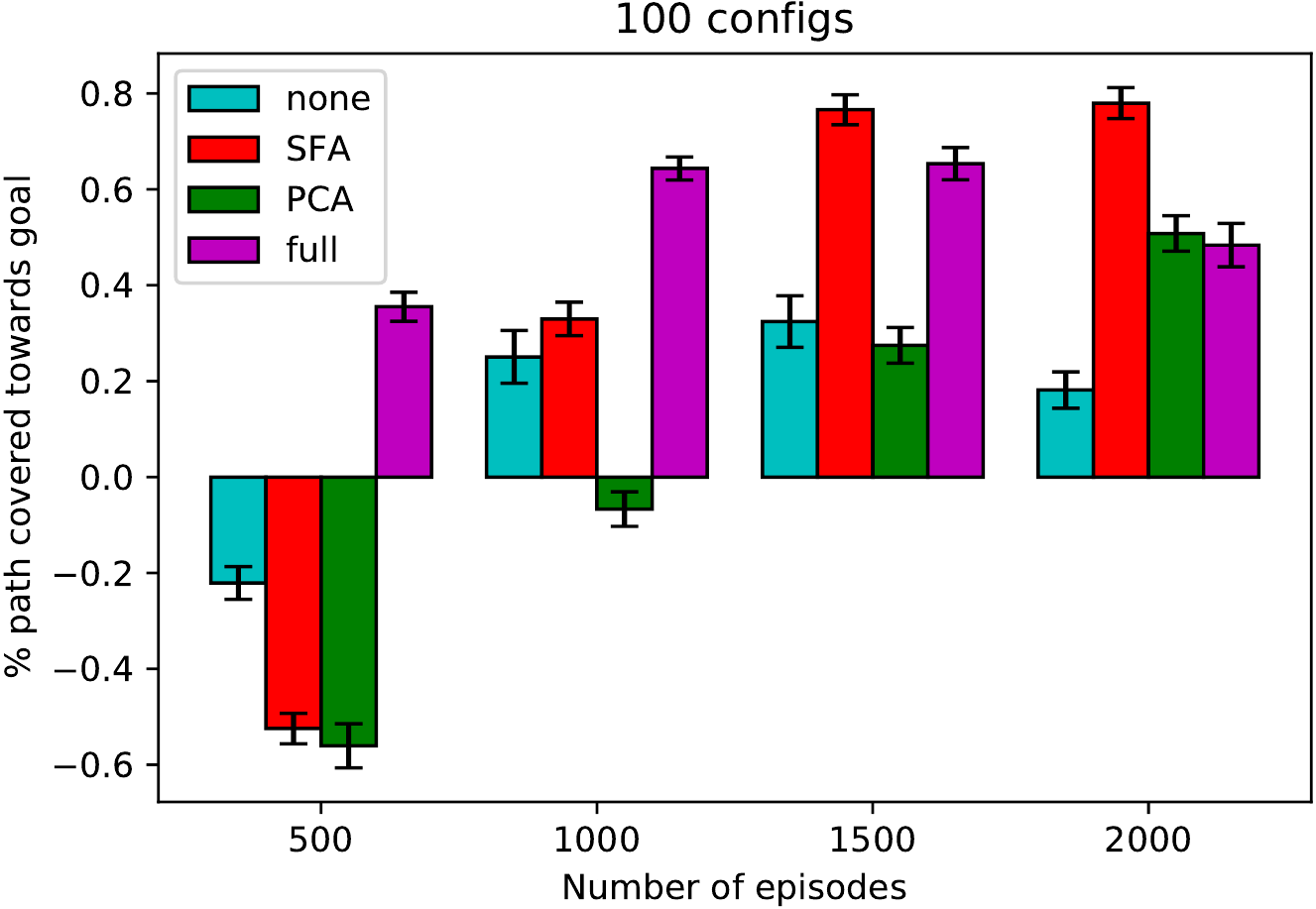}}
    \subfigure[200 configs \label{subfig:200config}]{\includegraphics[width=0.32\textwidth]{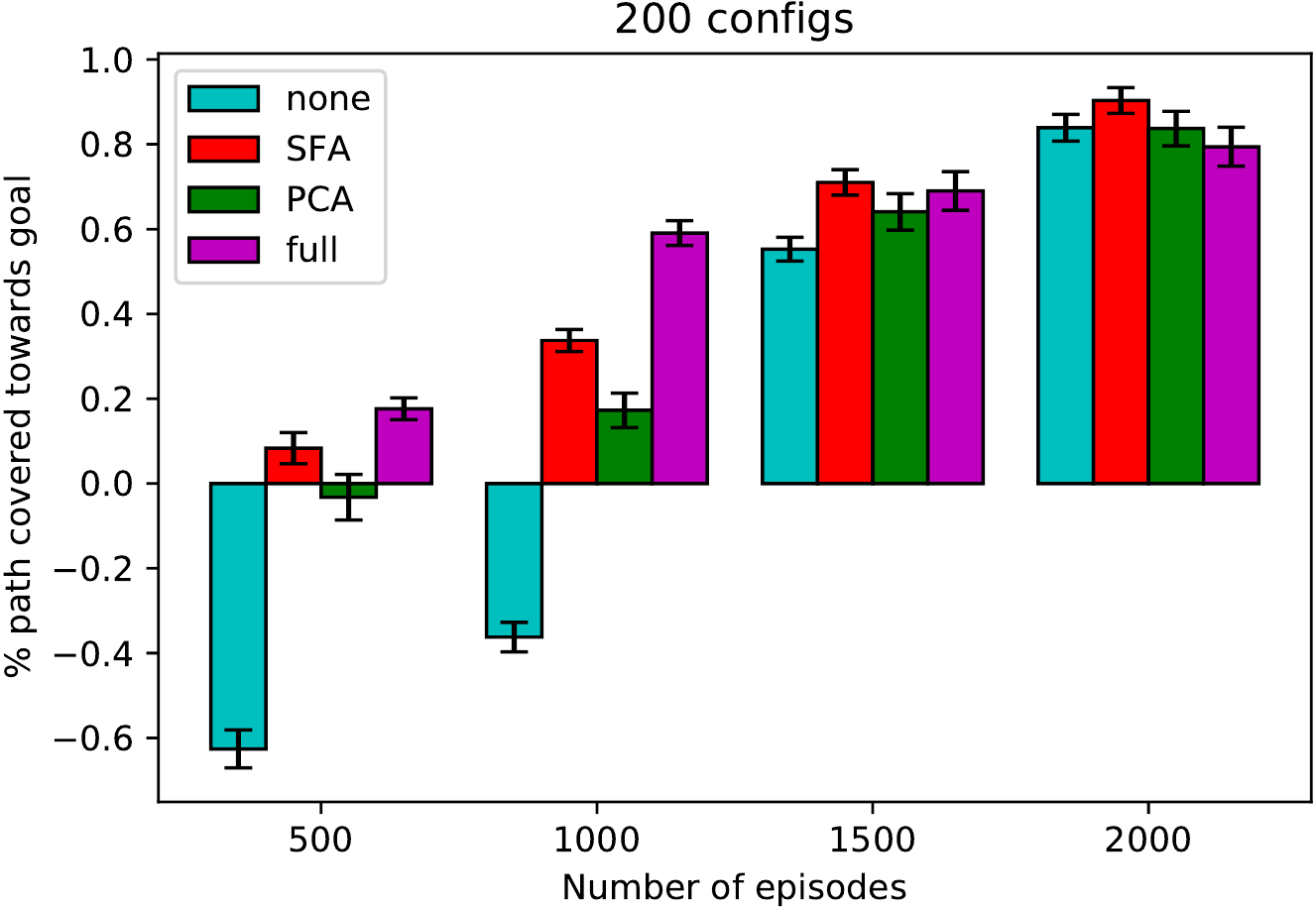}}
    \caption{Test performance vs. number of training episodes for different methods and configuration numbers.}
    \label{fig:epi_rewards}
\end{figure}

We also report the test results at different episodes in Figure \ref{fig:epi_rewards}. The figure suggests that using an additional set of features from a previously learned task helps generalization, especially when there are fewer number of configurations. We see that \texttt{transfer:sfa} performs on par with \texttt{transfer:none} in the initial stages of the training, then achieves the peak of its performance very quickly compared to other methods. On the other hand, the contribution of \texttt{transfer:full} steadily increases. This might be due to the distribution of information in the neurons. Only a few SFA units contain useful information for the task, and when these neurons are discovered with gradient descent in later steps, the performance increases rapidly. Moreover, \texttt{transfer:sfa} learns faster compared to other methods for less number of configurations (see 50 and 100 configs in Figure \ref{fig:epi_rewards}), and also requires less computation compared to \texttt{transfer:full} as there are less number of units, which is an essential property for small, autonomous systems with no access to a graphics processing unit (GPU).

\subsection{Correlation with High-Level Task Properties}
\label{subsec:correlation}

In this experiment, we investigate correlations between various units and high-level features. To this end, we freeze the network and run the policy for 100 episodes to collect hidden layer activations for PCA and SFA calculations. Then, we apply PCA and SFA with 100 components to the last hidden layer activations. To understand whether the transformed features capture any high-level features such as `distance from the tip to the goal' or joint angles that are helpful to solve the task, we calculate the correlation between those high-level features and PCA/SFA features. We also compare the correlations between the last layer activations to see the effect of the these transformations on the correlations.

\begin{figure}[htbp]
    \centering
    \subfigure[$D(x_{tip}, x_{goal})$]{\includegraphics[width=0.32\textwidth]{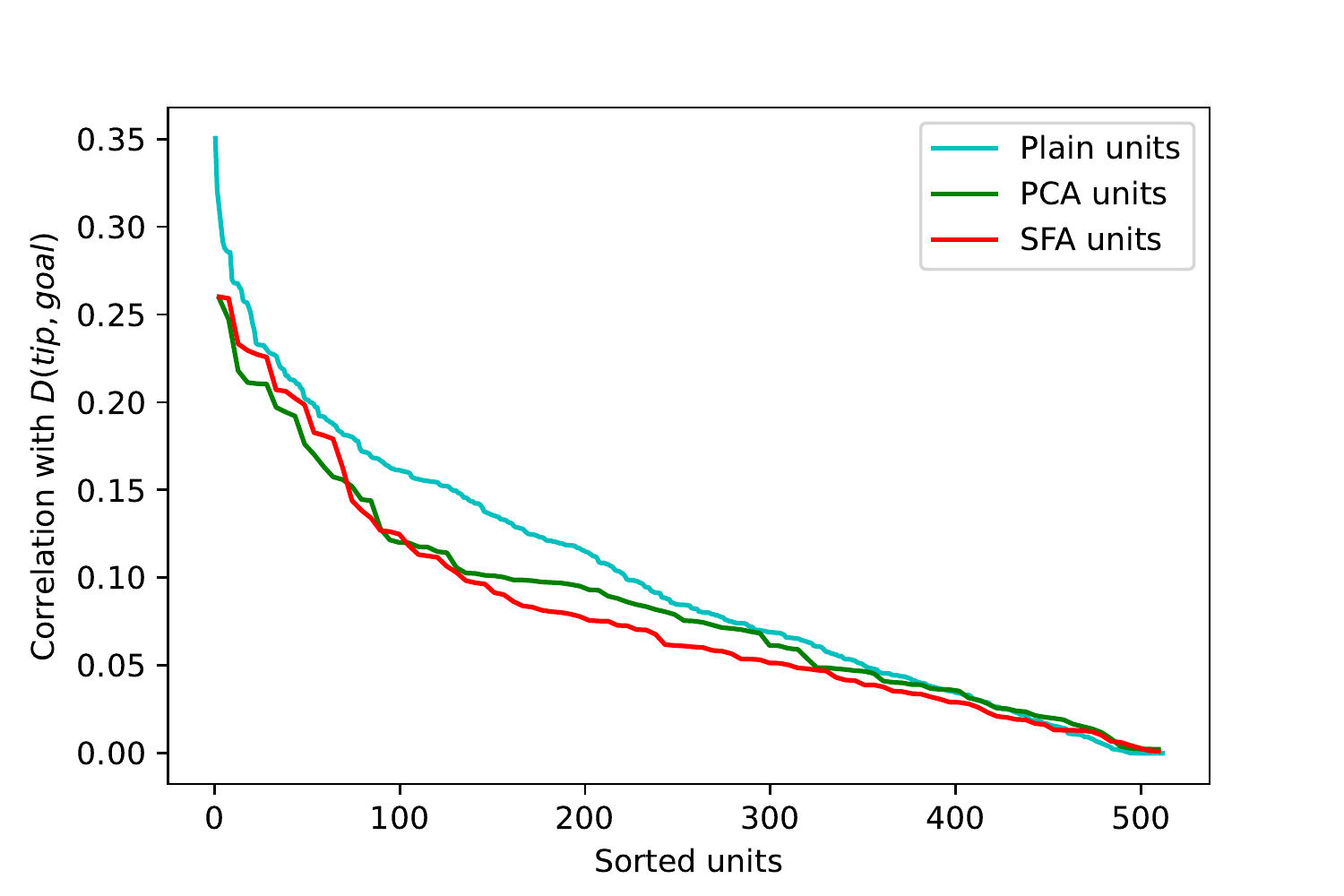}}
    \subfigure[$D(x_{tip}, x_{obstacle})$]{\includegraphics[width=0.32\textwidth]{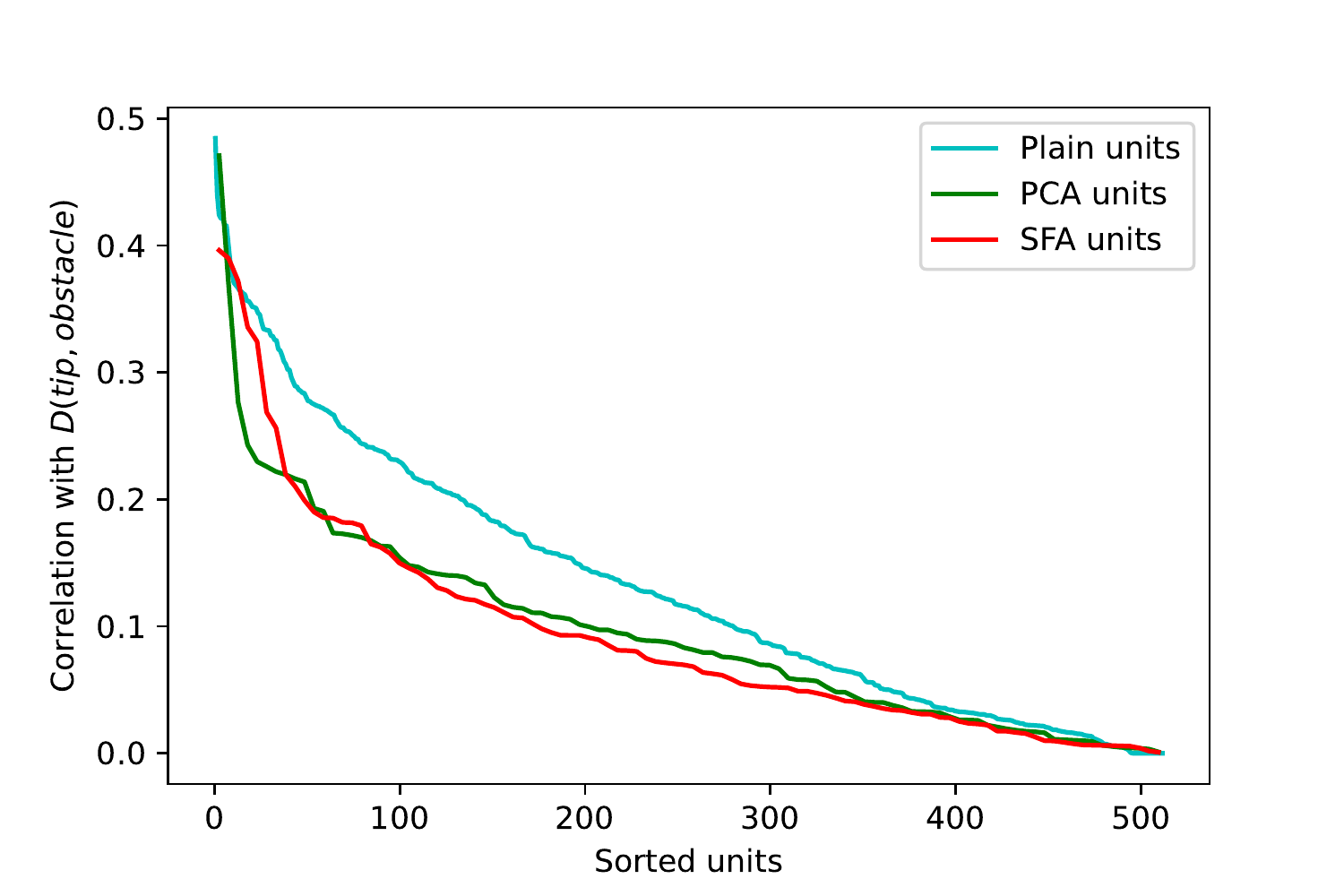}}
    \subfigure[`path blocked']{\includegraphics[width=0.32\textwidth]{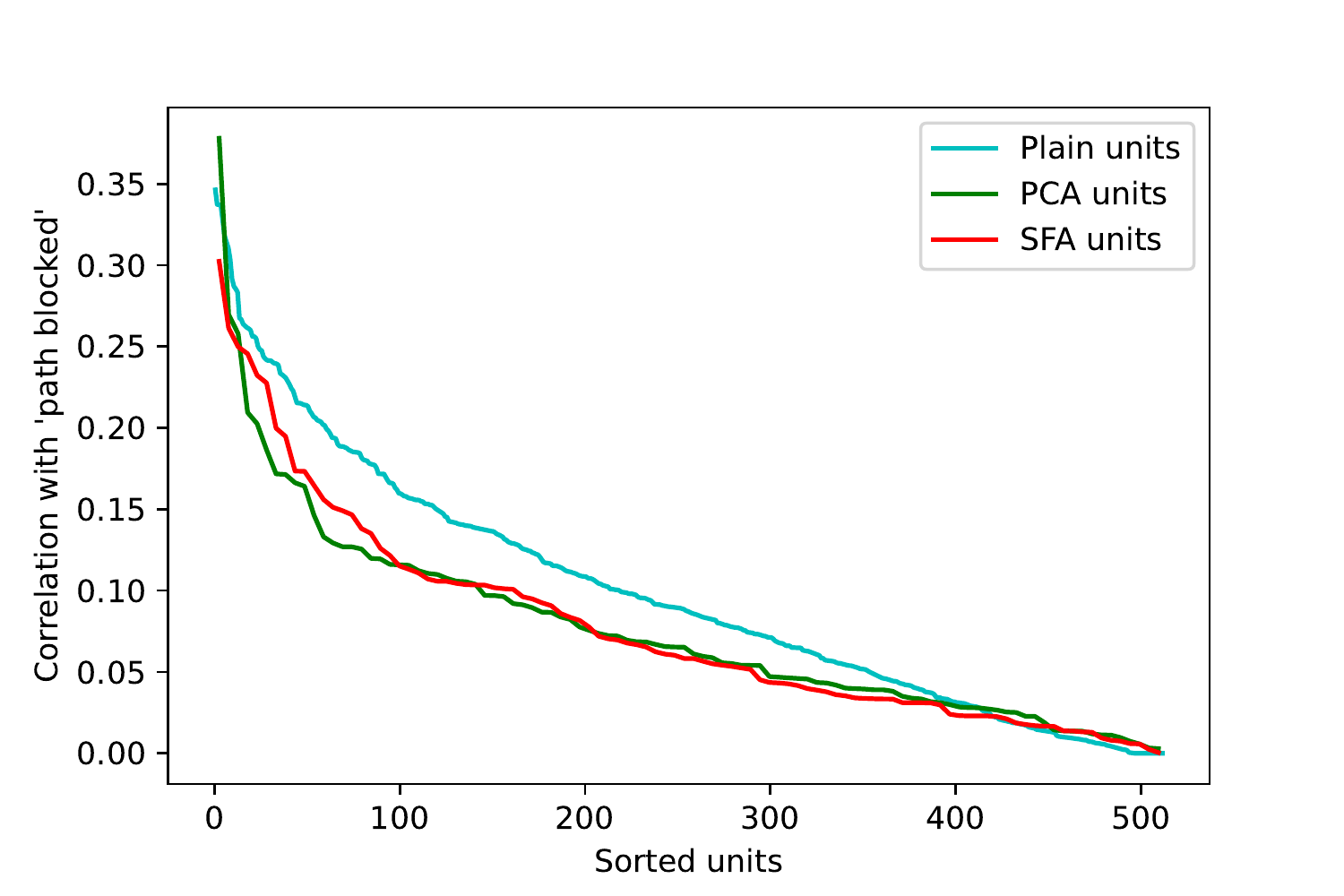}}
    
    \subfigure[Joint 0 \label{subfig:j0}]{\includegraphics[width=0.32\textwidth]{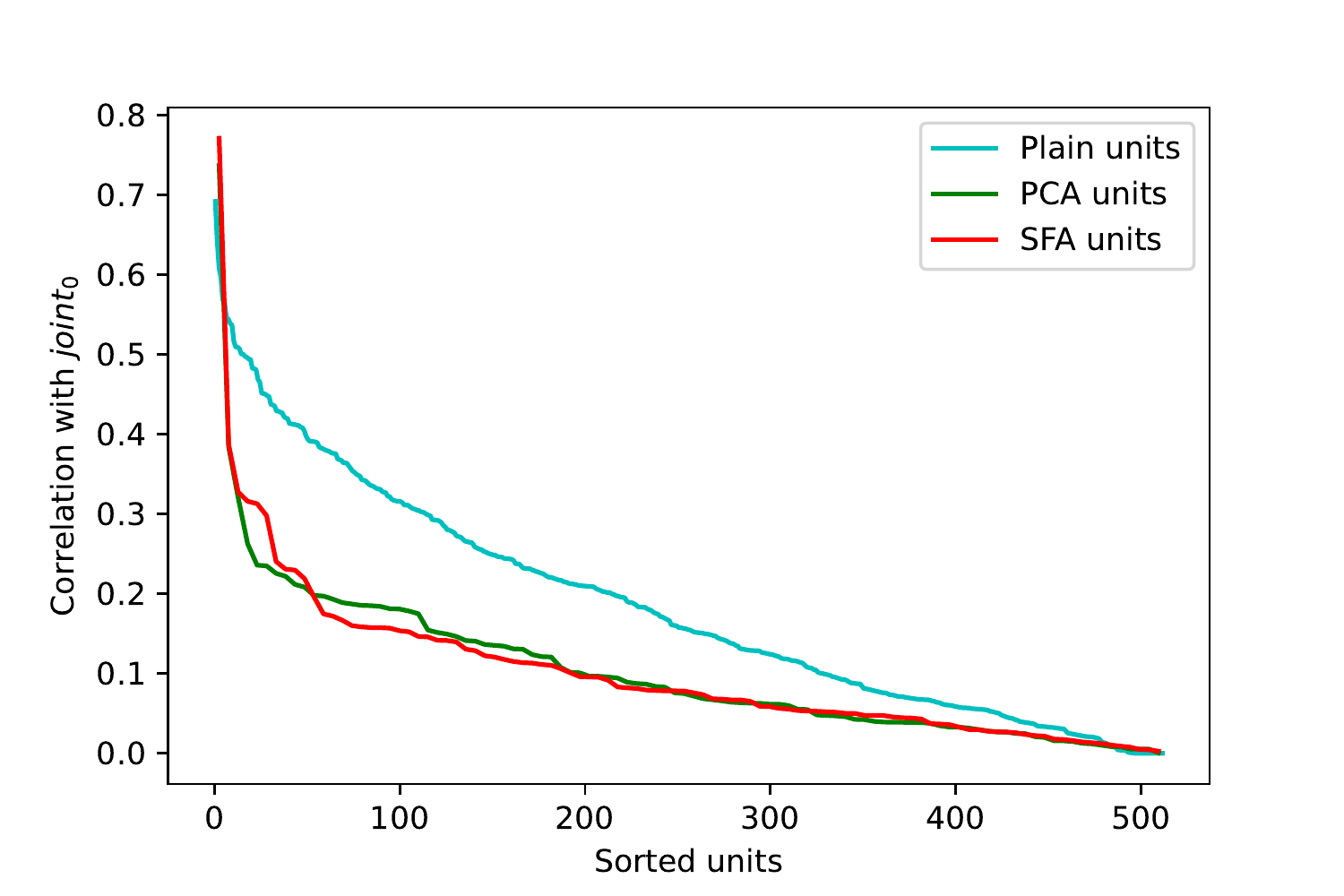}}
    \subfigure[Joint 1 \label{subfig:j1}]{\includegraphics[width=0.32\textwidth]{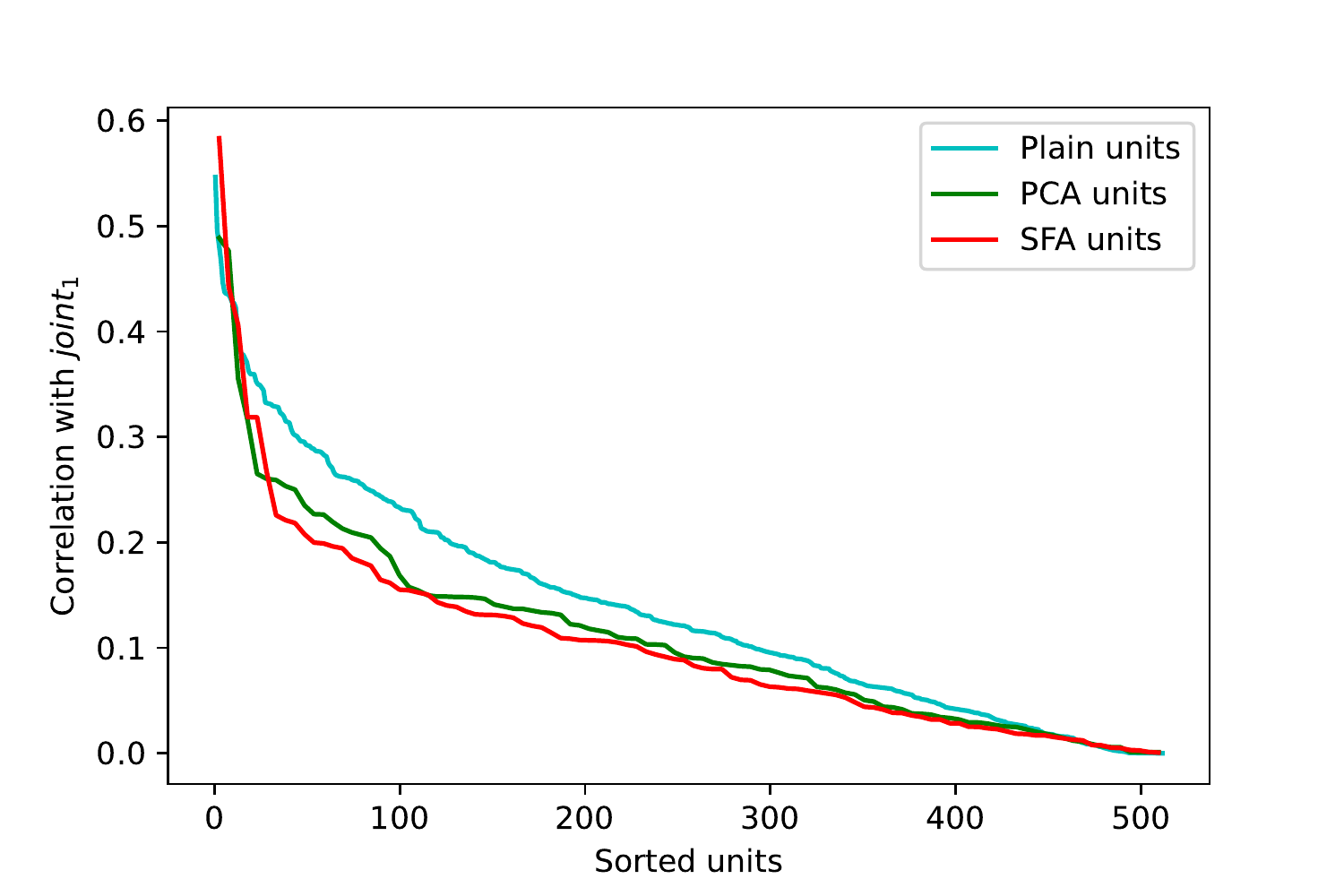}}
    \subfigure[Joint 2 \label{subfig:j2}]{\includegraphics[width=0.32\textwidth]{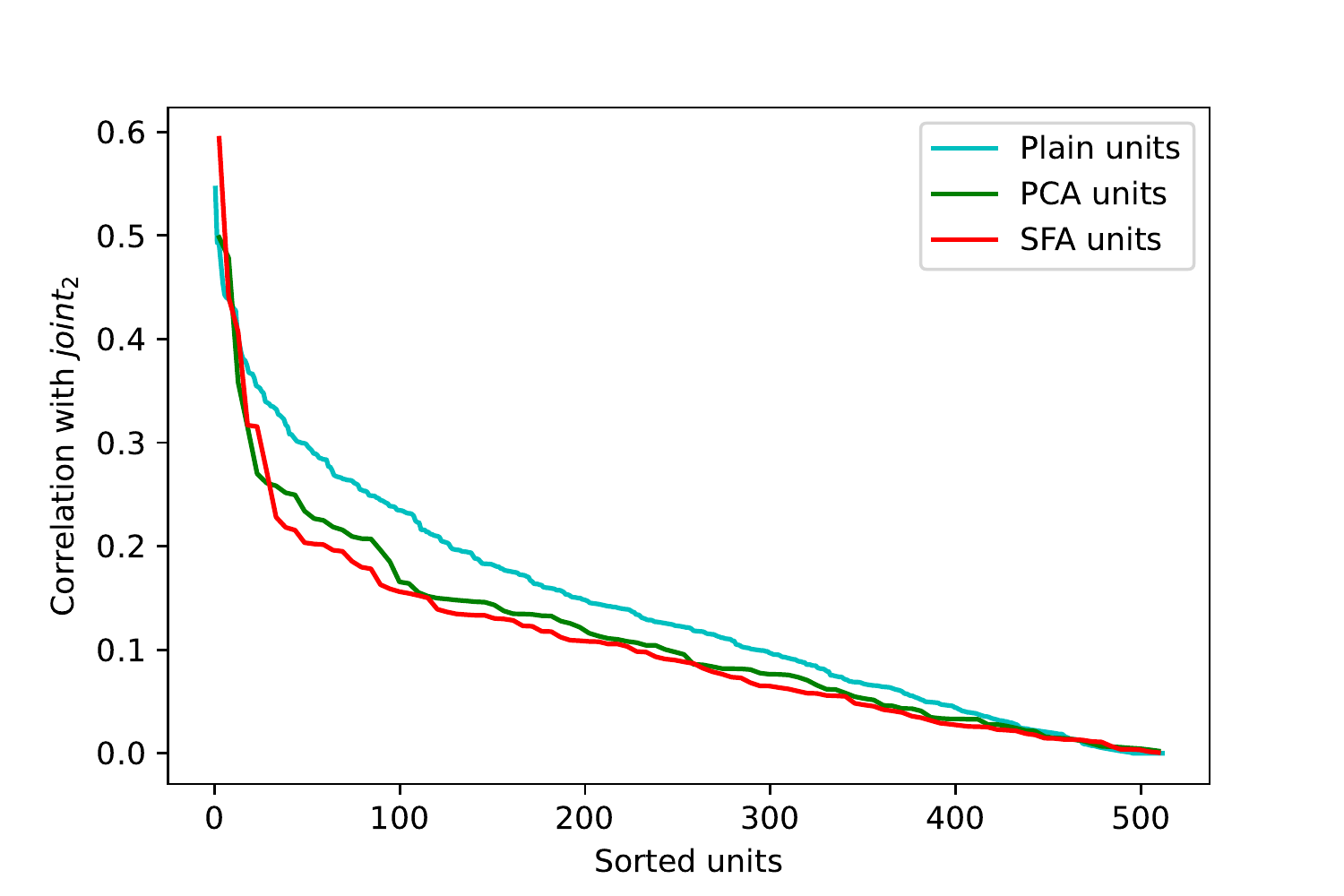}}

    \subfigure[Joint 3 \label{subfig:j3}]{\includegraphics[width=0.32\textwidth]{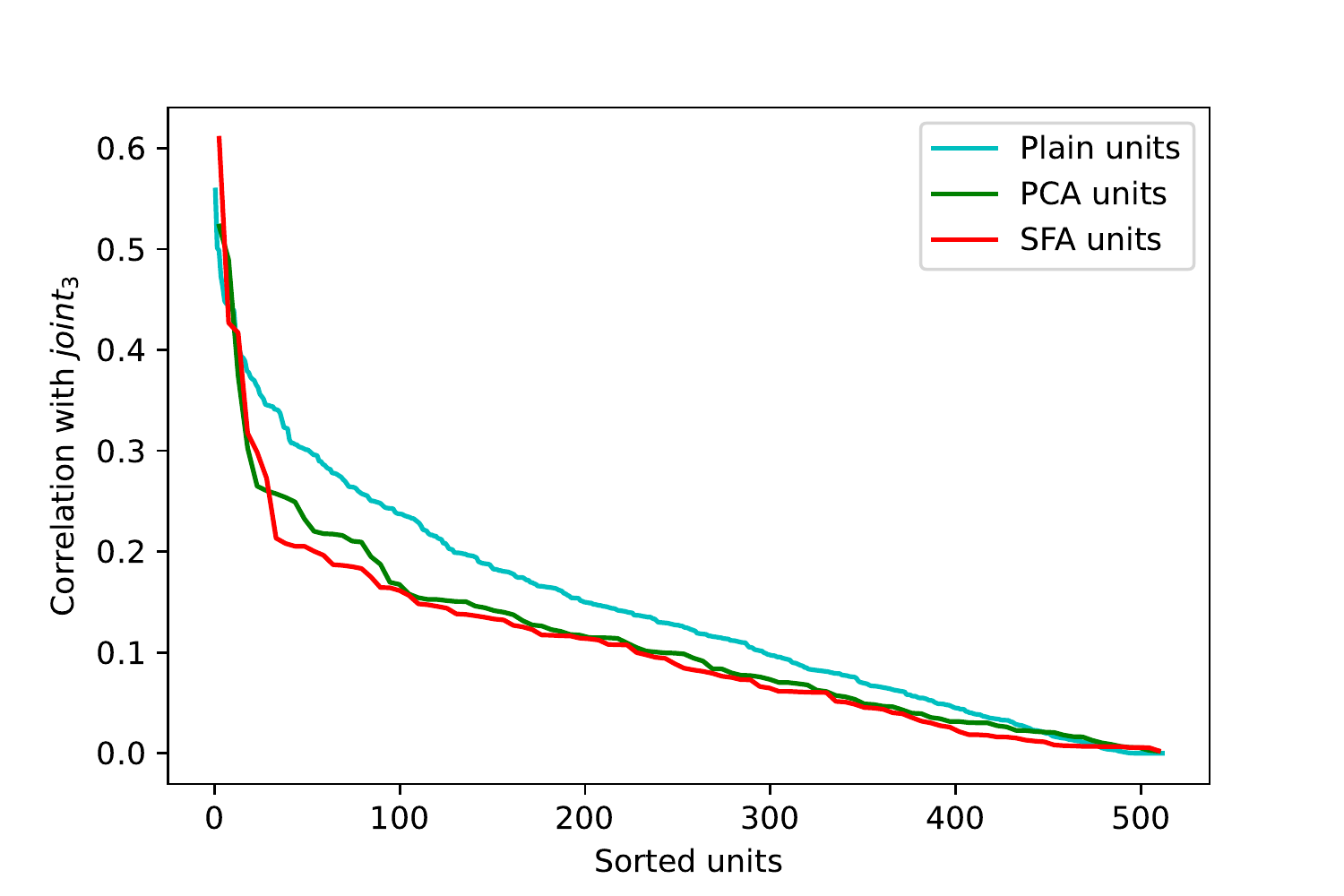}}
    \subfigure[Joint 4]{\includegraphics[width=0.32\textwidth]{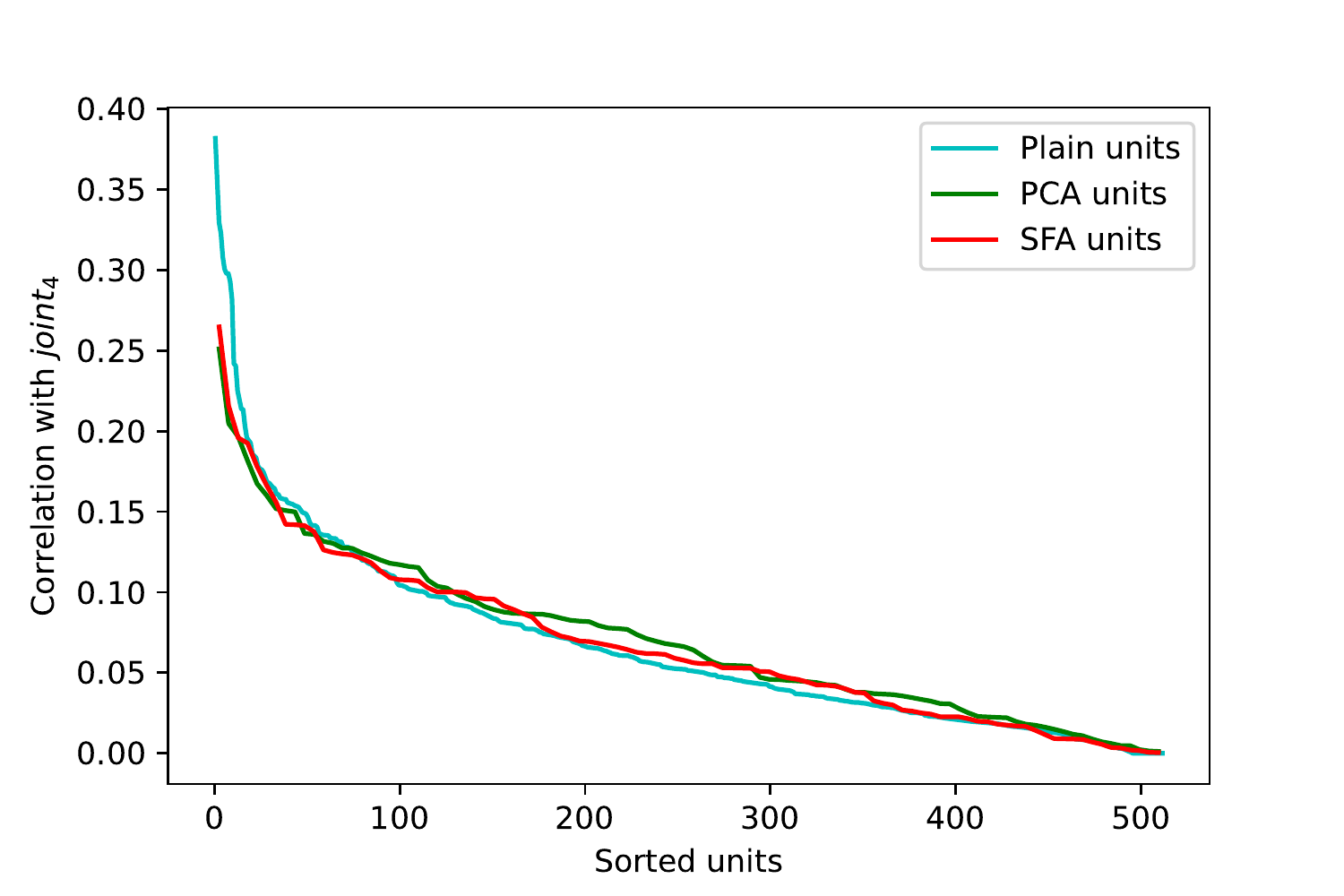}}
    \subfigure[Joint 5]{\includegraphics[width=0.32\textwidth]{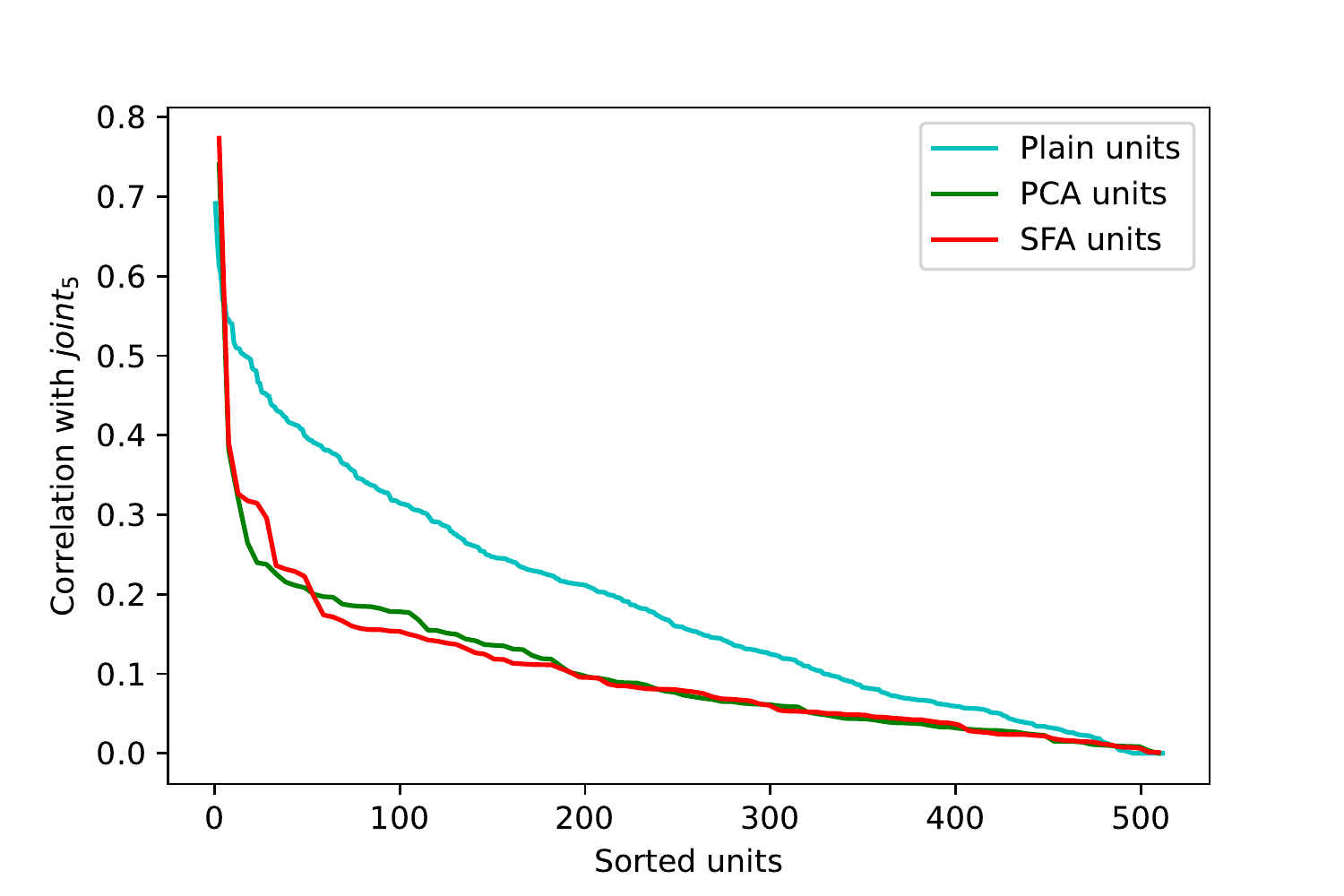}}

    \caption{Units are sorted by their correlation. SFA and PCA units are highlighted in red and green, respectively.}
    \label{fig:sort_transform}
\end{figure}

In Figure \ref{fig:sort_transform}, each subplot shows the distribution of correlations for each different high-level feature. Units are sorted by their correlation value for better visualization. After the transformation with PCA and SFA, most of the units become less correlated while only a few of them become highly correlated (e.g. Figures \ref{subfig:j0} and \ref{subfig:j1}). This is probably due to PCA and SFA objectives force components to be decorrelated. From Figure \ref{fig:sort_transform}, we can say that PCA and SFA capture the general neural response; they eliminate redundant neural responses and focus on the important ones by keeping the general response profile of the hidden layer. Out of the two, SFA unit distribution has more units with high correlation, especially for joint angles (Figures \ref{subfig:j0}-\ref{subfig:j3}). This might be one of the reasons that the learning performance of \texttt{transfer:sfa} is comparable to \texttt{transfer:full}, and even better for the case of less number of configurations since redundant and possibly noisy variations are filtered out by SFA.
\begin{table}[htbp]
    \tbl{Units that correlate the maximum with high-level features are reported. The numbers on the left and the right denotes the correlation and the index of the unit, respectively.}
    {\begin{tabular}{@{}cccc}
    \toprule
         High-level feature & Full activations & PCA features & SFA features \\
         \colrule
         Joint 1 & 0.69 / 297 & 0.74 / 2 & \textbf{0.77} / 1 \\
         Joint 2 & 0.55 / 249 & 0.49 / 5 & \textbf{0.58} / 3 \\
         Joint 3 & 0.55 / 249 & 0.50 / 5 & \textbf{0.59} / 3 \\
         Joint 4 & 0.56 / 249 & 0.52 / 5 & \textbf{0.61} / 3 \\
         Joint 5 & \textbf{0.38} / 471 & 0.20 / 3 & 0.19 / 2 \\
         Joint 6 & 0.69 / 297 & 0.74 / 2 & \textbf{0.77} / 1 \\
         $D(x_{\text{tip}}, x_{\text{goal}})$ & \textbf{0.35} / 303 & 0.26 / 2 & 0.26 / 2 \\
         $D(x_{\text{tip}}, x_{\text{obstacle}})$ & \textbf{0.48} / 147 & 0.47 / 3 & 0.40 / 3 \\
         `path blocked' & 0.35 / 147 & \textbf{0.38} / 3 & 0.25 / 2 \\
    \botrule
    \end{tabular}}
    \label{tab:correlations}
\end{table}

In Table \ref{tab:correlations}, we select the most correlated units from each method and report their correlations. We see that all methods have high correlation with joint angles and slightly lower correlation with high-level features that relates the tip position to other objects. The high correlation with joint angles is an expected result as the agent should know about the location of the arm in order to navigate it to the goal position.

\subsection{Visualizing Features}
\label{subsec:visualization}
\begin{figure}[htbp]
    \centering
    \subfigure[SFA responses to the tip position. \label{subfig:sfatip20}]{\includegraphics[width=0.9\textwidth]{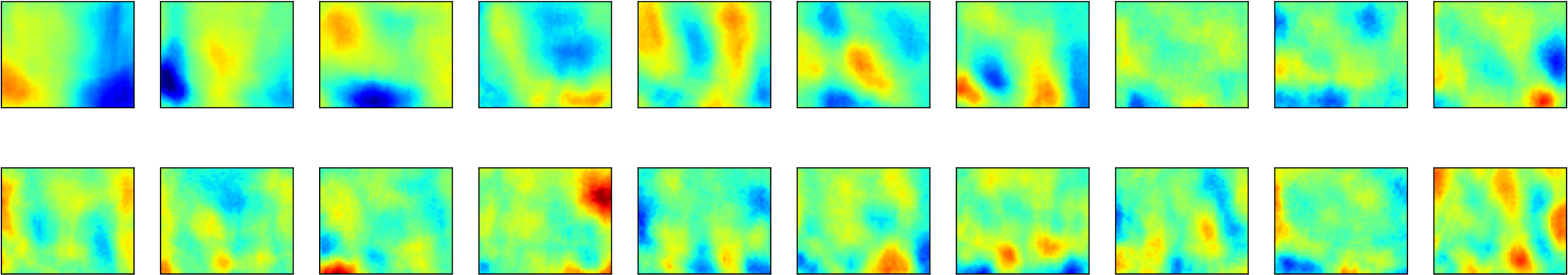}}
    \subfigure[PCA responses to the tip position. \label{subfig:pcatip20}]{\includegraphics[width=0.9\textwidth]{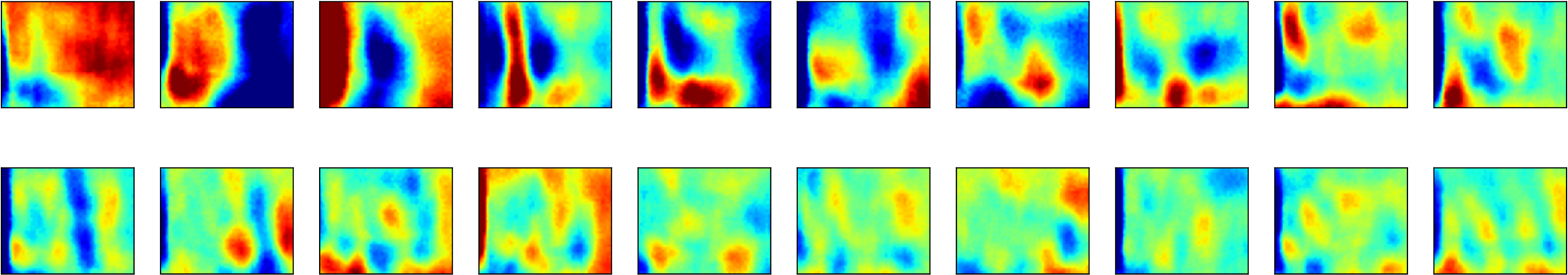}}
    \caption{First 20 SFA and PCA units' responses to different tip positions.}
    \label{fig:tip20}
\end{figure}

We created heatmaps for the responses of different SFA and PCA units to varying inputs. In Figure \ref{fig:tip20}, each rectangle represents a unit's average response to different tip locations. For example, for the top left unit in Figure \ref{subfig:sfatip20}, the response is high when the tip is located around the bottom left corner of the table, and it is low around the bottom right corner. To create these figures, we averaged out over other variables (i.e. the goal position and the obstacle position). We see that both SFA and PCA units response to blob-like regions of the table. The first few units are very compact and can be partially treated as symbolic representations (e.g. in Figure \ref{subfig:sfatip20}, the first and the third units detect the position in $x$-axis and $y$-axis, respectively.) As the unit number increases, these regions start to become scattered.

\begin{figure}[htbp]
    \centering
    \subfigure[SFA responses to the relative goal position. \label{subfig:sfarg}]{\includegraphics[width=0.9\textwidth]{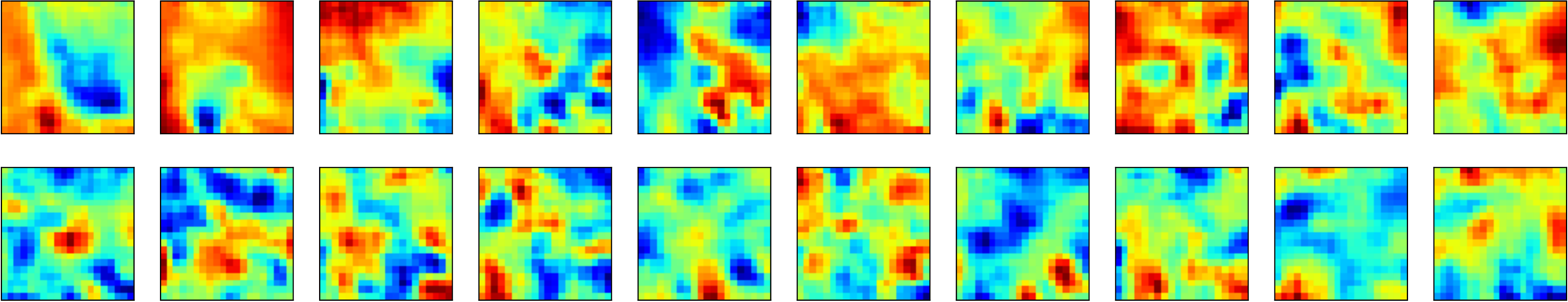}}
    \subfigure[PCA responses to the relative goal position. \label{subfig:pcarg}]{\includegraphics[width=0.9\textwidth]{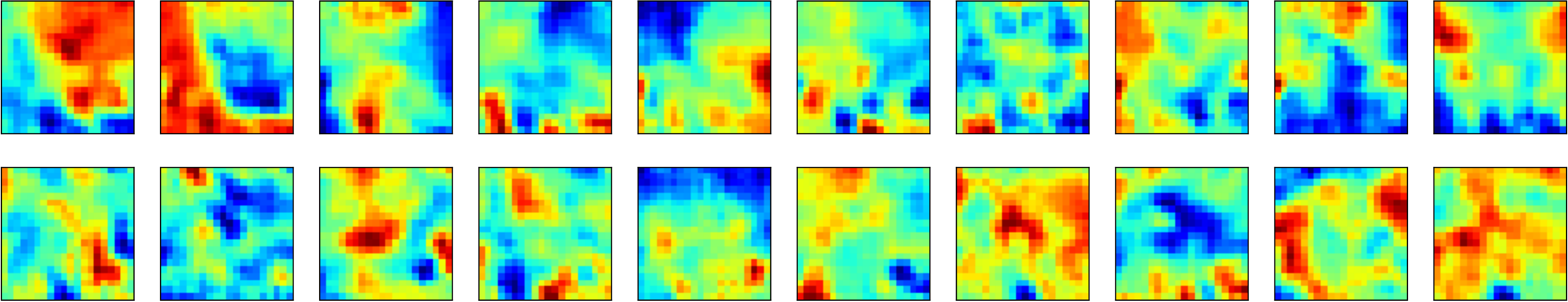}}
    \caption{First 20 SFA and PCA features' responses to the relative goal position. The center point represents (0, 0), i.e. zero distance between the tip and the goal.}
    \label{fig:relgoal}
\end{figure}

\begin{figure}[htbp]
    \centering
    \subfigure[SFA responses to the relative obstacle position.\label{subfig:sfaro}]{\includegraphics[width=0.9\textwidth]{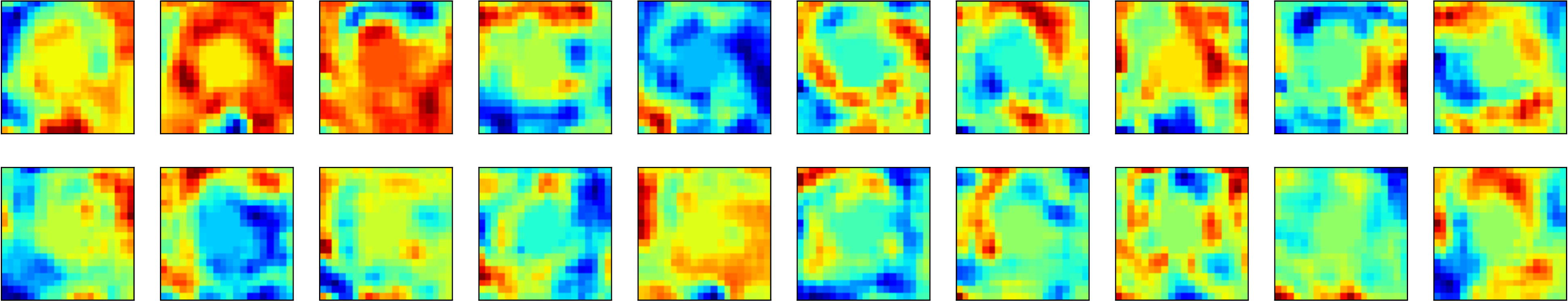}}
    \subfigure[PCA responses to the relative obstacle position. \label{subfig:pcaro}]{\includegraphics[width=0.9\textwidth]{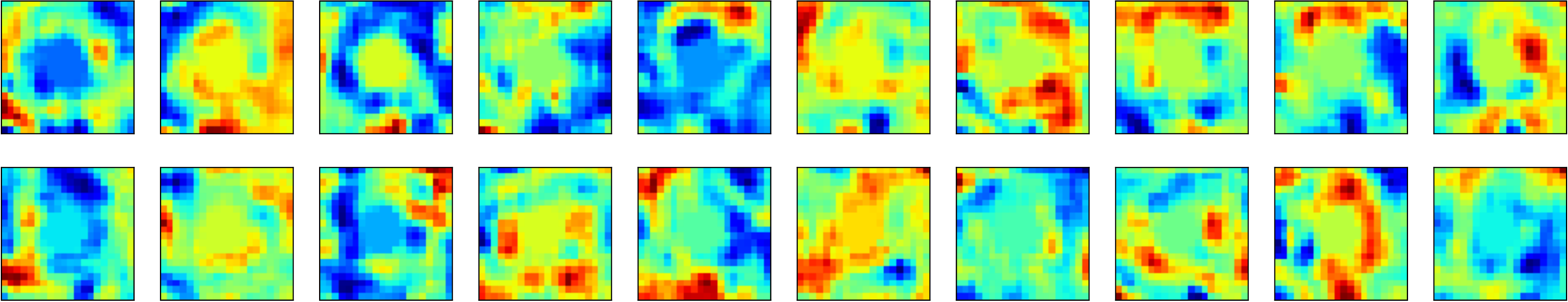}}
    \caption{First 20 SFA and PCA features' responses to the relative obstacle position. The center point represents (0, 0), i.e. zero distance between the tip and the obstacle.}
    \label{fig:relobs}
\end{figure}

While the tip position is a useful information to solve the task, the agent should also know the relative position of the goal and the obstacle with respect to its tip to successfully navigate in the environment. To understand the responses for relative distance to the goal and the obstacle, we created heatmaps with the similar procedure in Figures \ref{fig:relgoal} and \ref{fig:relobs}. The center point of each subplot represents the zero distance (i.e. $x_{\text{goal}}-x_{\text{tip}}=(0, 0)$). In Figure \ref{fig:relgoal}, most of the units are scattered except the first few ones. This figure suggests that both SFA and PCA are not very successful at covering the relative goal position compactly; the information is distributed into many units as in plain activations. However, in Figure \ref{fig:relobs}, we see that both methods generates neuron responses that are inactive when the tip is close to the obstacle (e.g. the second SFA unit and the first PCA unit). The response of SFA is more uniformly distributed compared to PCA which might be one of the reasons for higher transfer performance when an L-shaped obstacle is introduced.

{}

\begin{figure}[htbp]
    \centering
    \subfigure[303rd neuron \label{subfig:relgoal}]{\includegraphics[width=0.2\textwidth]{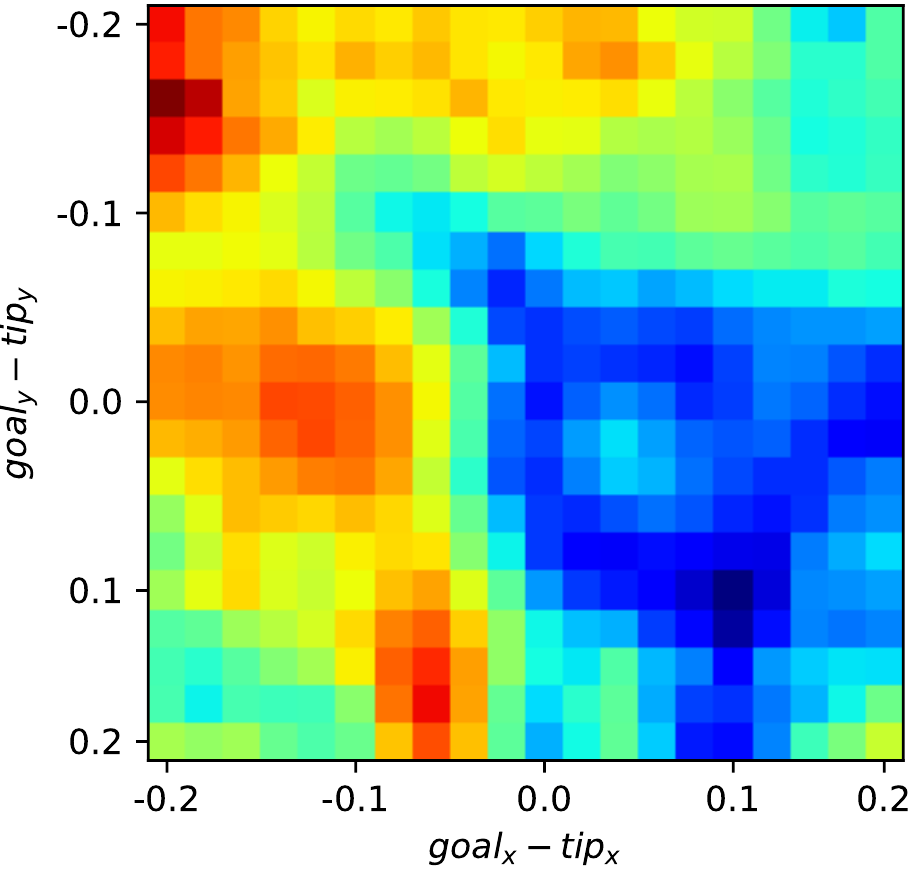}}
    \subfigure[147th neuron \label{subfig:relobs}]{\includegraphics[width=0.2\textwidth]{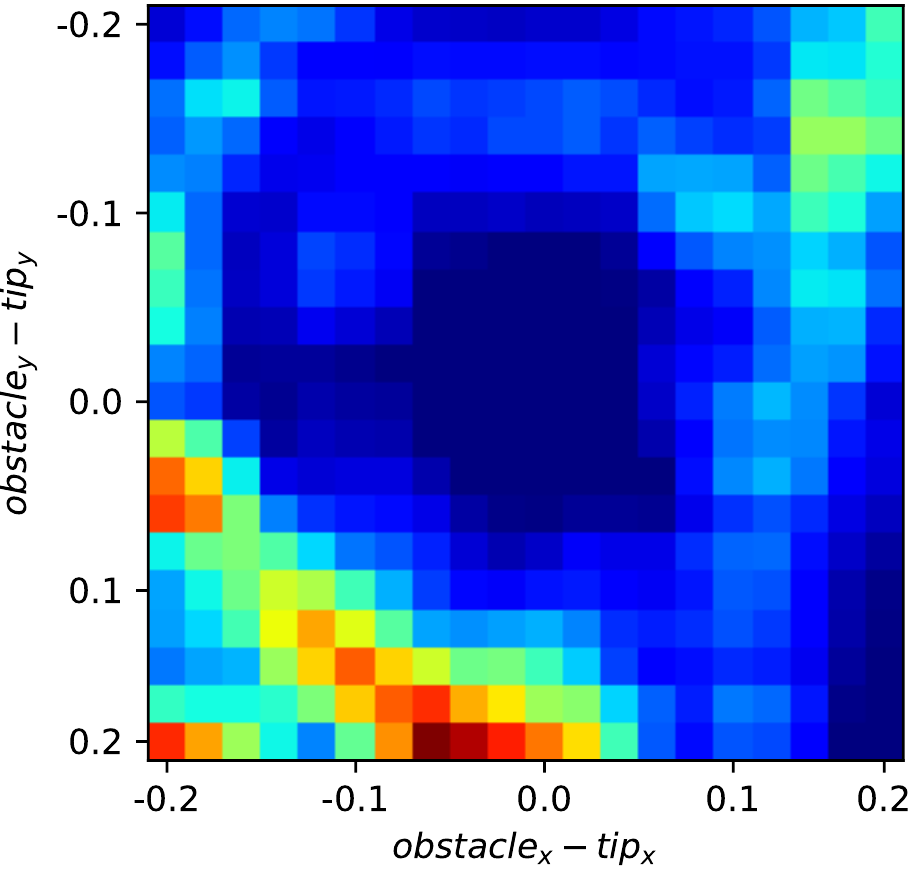}}
    \caption{Neurons that correlate the most with \ref{subfig:relgoal} the relative goal position and \ref{subfig:relobs} the relative obstacle position.}
    \label{fig:relneurons}
\end{figure}

To understand whether these two high-level information are present in the network prior to transformation, we visualized the neurons that correlate the most with the relative goal position and the relative obstacle position in Figure \ref{fig:relneurons}. For the bottom half in Figure \ref{subfig:relgoal}, we see that the most correlating neuron partially responds to $x$-axis location of the relative goal position. On the other hand, in Figure \ref{subfig:relobs}, the response is similar to SFA and PCA units; it becomes inactive when distance is below some threshold.

We conclude that even though the plain activations carry a distributed representation in general, they might represent some high-level information as a by-product which can be further processed and refined. We see that SFA and PCA transformations create high-level features to detect the tip position, however, this is not as compact as for the relative positions. Note that both SFA and PCA are unsupervised methods, not specifically tailored for creating more symbolic information. However, since they automatically create low-complexity signals in a principled way, they are of good candidates of a more complex system.

\section{Conclusion}
\label{sec:conclusion}
As the literature expands rapidly, the integration of architectures that are optimized for a specific scenario will gain more importance. In this work, we provide a principled way for transferring the existing knowledge in a network to new problems. Our experimental results show that applying SFA, an unsupervised method, to the last hidden layer of a trained network generates features that are useful for skill transfer. Transfer with SFA performs better with less number of units compared to transfer with the full layer. This is a desirable property for life-long learning systems because of its resource economy. We see that features that are generated from SFA are more interpretable and they can be treated as quasi-symbolic information. Moreover, due to its formulation, the components with lowest eigenvalue will contain the most useful and symbol-like information. Therefore, one can always tune the number of units by simply picking the first $k$ components with the lowest eigenvalues, a procedure familiar to PCA.

As a future work, we plan to extend the experimental setup to a more continual, open-ended environment so that we can learn representations that are of progressive complexity to see the limits of the method. Another possible future work is to integrate the slowness prior to the objective of learning to make the whole system an end-to-end architecture. The effectiveness of other dimensionality reduction methods such as independent component analysis for skill transfer can be also explored as a future work.

\section*{Acknowledgement}
This  research  was  supported by the International Joint Research Promotion Program of Osaka University  under the project “Developmentally and biologically realistic modeling of perspective invariant action understanding” and T\"{U}B\.{I}TAK (The Scientific and Technological Research Council of Turkey) ARDEB 1001 program (project number: 120E274).

\bibliographystyle{tADR}
\bibliography{ref}

\end{document}